\documentclass[conference,10pt]{IEEEtran} 

\IEEEoverridecommandlockouts

\usepackage[english]{babel}
\usepackage{ifpdf}
\usepackage{cite} 
\usepackage{url}
\ifpdf
	\usepackage[pdftex]{graphicx}
	\graphicspath{{./figures/}}
\else
	\usepackage[dvips]{graphicx}
	\graphicspath{./figures/}
\fi
\usepackage{color}
\usepackage{pgf, tikz, pgfplots}
\usetikzlibrary{shapes, arrows, automata}
\usepackage{caption} 
\usepackage{subcaption} 
\usepackage{amsmath}
\usepackage{amsfonts, amssymb, amsthm}
\usepackage{mathrsfs}
\usepackage{upgreek}
\usepackage{algorithm,algpseudocode}
\usepackage{enumerate}
\usepackage{multirow}
\usepackage{rotating}
\usepackage{needspace}

\input{mysymbol.sty}

\def\intersection{-0.965mm}
\def\intereq{-0.965mm}

\newcommand{\QED}{\hfill\ensuremath{\blacksquare}}

\def\Tr{\mathsf{T}}

\title{MIMO Graph Filters for Convolutional\\Neural Networks}
\author{Fer\hspace{0.02cm}nando~Gama,~
        Antonio~G.~Marques,~
        Alejandro~Ribeiro,~
        and~Geert~Leus
\thanks{Work supported by USA NSF CCF-1717120 and ARO W911NF1710438, Spanish MINECO grants No. TEC2013-41604-R and TEC2016-75361-R. F. Gama and A. Ribeiro are with the Dept. of Electrical and Systems Eng., Univ. of Pennsylvania., A. G. Marques is with the Dept. of Signal Theory and Comms., King Juan Carlos Univ., G. Leus is with the Dept. of Microelectronics, Delft Univ. of Technology.  Emails: \{fgama,aribeiro\}@seas.upenn.edu, antonio.garcia.marques@urjc.es, and g.j.t.leus@tudelft.nl
.
}
}

\begin{document}

\maketitle
\begin{abstract}
Superior performance and ease of implementation have fostered the adoption of Convolutional Neural Networks (CNNs) for a wide array of inference and reconstruction tasks. CNNs implement three basic blocks: convolution, pooling and pointwise nonlinearity. Since the two first operations are well-defined only on regular-structured data such as audio or images, application of CNNs to contemporary datasets where the information is defined in irregular domains is challenging. This paper investigates CNNs architectures to operate on signals whose support can be modeled using a graph. Architectures that replace the regular convolution with a  so-called linear shift-invariant graph filter have been recently proposed. This paper goes one step further and, under the framework of multiple-input multiple-output (MIMO) graph filters, imposes additional structure on the adopted graph filters, to obtain three new (more parsimonious) architectures. The proposed architectures result in a lower number of model parameters, reducing the computational complexity, facilitating the training, and mitigating the risk of overfitting. Simulations show that the proposed simpler architectures achieve similar performance as more complex models.
\end{abstract}
\begin{IEEEkeywords}
Convolutional neural networks, network data, graph signal processing, MIMO.
\end{IEEEkeywords}

\vspace{\intersection}

\section{Introduction} \label{sec_intro}

Convolutional Neural Networks (CNNs) have emerged as the information processing architecture of choice in a wide range of fields as diverse as pattern recognition, computer vision and medicine, for solving problems involving inference and reconstruction tasks \cite{bruna13-scattering, lecun10-vision, greenspan16-medical}. CNNs have demonstrated remarkable performance, as well as ease of implementation and low online computational complexity \cite{lecun15-deeplearning, najafabadi15-cnnbigdata}. CNNs take the input data and process it through several layers, each of which performs three simple operations on the output of the previous layer. These three operations are convolution, pointwise nonlinearity and pooling. The objective of such an architecture is to progressively extract useful information, from local features to more global aspects of the data. This is mainly achieved by the combination of convolution and pooling operations which sequentially combine data that is located further away. The nonlinearity dons the architecture with enough flexibility to represent a richer class of functions that may describe the problem.

One of the most outstanding characteristics of CNNs is that the filters used for convolution can be efficiently \emph{learned} from training datasets by means of a backpropagation algorithm \cite{rumelhart86-backprop}. This implies that the CNN architecture is capable of \emph{learning} which are the most useful features for the task at hand. Intimately related to the capability of successful training, is the fact that the filters used are small, thus containing few parameters, making training easier.
While, convolution and pooling are well-defined only in  regular domains such as time or space, contemporary data is increasingly being described on domains that exhibit more irregular behavior \cite{lazer09-compsoc}, with examples including marketing, social networks, or genetics \cite{jackson08-social, davidson02-genetics, haupt08-csnetwork}. With the objective of extending the remarkable performance of CNNs to broader data domains, extensions capable of processing network data have been developed \cite{bruna14-deepspectralnetworks, henaff15-deepgraph, niepert16-learningcnn, defferrard17-cnngraphs, kipf17-classifgcnn, gama17-nvgf, pasdeloup17-approxtrans, du17-topoadapt}, see \cite{bronstein17-geomdeeplearn} for a survey. In particular, the works of \cite{bruna14-deepspectralnetworks, defferrard17-cnngraphs} make use of the concept of graph filters (GFs) \cite{sandryhaila14-freq, segarra17-linear} to extend the convolution operation to graph signals \cite{sandryhaila13-dspg
}. Leveraging the framework of multiple-input multiple-output (MIMO) GFs on existing results, this paper proposes three novel architectures for GF-based CNNs. The main idea is to replace the bank of parallel GFs with a more structured filtering block which reduces the degrees of freedom (number of parameters) on each layer. This new architecture facilitates the training, incurs reduced computational complexity, and can be beneficial to avoid overfitting.

Section~\ref{sec_gcnn} introduces notation and reviews existing GF-based CNNs under the framework of MIMO GFs. Section~\ref{sec_simple} describes the novel architectures. Section~\ref{sec_sims} presents simulations showing the benefits of our schemes. Concluding remarks are provided in Section~\ref{sec_conclusions}.  

\vspace{\intersection}

\section{CNNs on Graph Signals} \label{sec_gcnn}

Let $\bbx \in \ccalX$ be the input data defined on some field $\ccalX$ and let $\bby \in \ccalY$ be the output data such that $\bby = f (\bbx)$ for some (unknown) function $f: \ccalX \to \ccalY$. The general objective in machine learning is to estimate or \emph{learn} the function $f$ \cite{kearns94-introlearning}.

A neural network is an information processing architecture that aims at constructing an estimator $\hhatf$ that consists of a concatenation of layers, each of which applies three simple operations on the output of the layer before, namely a linear transform, a pointwise nonlinearity and a pooling operator. Formally, the estimator $\hhatf$ can be written as $\hhatf = f_{L} \circ f_{L-1} \circ \cdots \circ f_{1}$ where $f_{\ell}$ denotes the operations to be applied at layer $\ell=1,\ldots,L$ \cite{goodfellow16-deeplearn}. Denote by $\bbx_{\ell} \in \ccalX_{\ell}$ the $N_{\ell}$-dimensional output of layer $\ell$ defined over field $\ccalX_{\ell}$, by $\bbA_{\ell}: \ccalX_{\ell-1} \to \ccalX'_{\ell}$ a linear transform between fields $\ccalX_{\ell-1}$ and $\ccalX'_{\ell}$, by $\sigma_{\ell}:\ccalX'_{\ell} \to \ccalX'_{\ell}$ a pointwise nonlinearity, and by $\ccalP_{\ell}: \ccalX'_{\ell} \to \ccalX_{\ell}$ the pooling operator. Then, each layer can be described as $\bbx_{\ell} = f_{\ell}(\bbx_{\ell-1}) = \ccalP_{\ell} \{ \sigma_{\ell} ( \bbA_{\ell} \bbx_{\ell-1}) \}$, $\ell=1,\ldots,L$ with $\ccalX_{0} \equiv \ccalX$ the input data field and $\ccalX_{L} \equiv \ccalY$ the output data field.

In particular, a CNN assumes that the linear operator $\bbA_{\ell}$ is comprised of a collection of $F_{\ell}$ filters of small support $K_{\ell}$, $\bbA_{\ell} = \{\bbh_{\ell,1},\ldots,\bbh_{\ell,F_{\ell}}\}$. Then, the application of the linear operator yields a collection of signals $\{\bbh_{\ell,1} \ast \bbx_{\ell-1}, \ldots, \bbh_{\ell,F_{\ell}} \ast \bbx_{\ell-1}\}$, in which each element $i = 1,...,N_{\ell-1}$,
\begin{equation} \label{eqn_regular_conv}
\vspace{\intereq}
[\bbh_{\ell,f} \ast \bbx_{\ell-1}]_{i} 
	= \sum_{k=0}^{K_{\ell}-1} [\bbh_{\ell,f}]_{k} [\bbx_{\ell-1}]_{i-k}
\vspace{\intereq}
\end{equation}
is considered a feature, $f=1,\ldots,F_{\ell}$, and where we assumed that $[\bbx_{\ell-1}]_{i-k}=0$ for $i \leq k$. Using filters with a small support has a twofold goal. First, the convolution operation linearly relates \emph{nearby} values (consecutive time instants, or neighboring pixels) and consolidates them in a feature value that aggregates this local information. Second, such filters only have a few parameters and, therefore, are easy to be \emph{learned} from data. We also note that the pooling operation serves the function of changing the \emph{resolution} of data, so that on each layer, the \emph{nearby} values that are related by the convolution operator are actually located further away. In this way, the convolution and pooling operations act in tandem to guarantee that the CNN aggregates information at different levels, from local to global.

The operation of convolution, in particular, depends upon the existence of a notion of neighborhood. Such a notion also exists in domains like manifolds and graphs, and thus, the convolution can be extended to operate on signals defined on these irregular domains. In particular, for signals defined on graphs, let us start by considering the graph $\ccalG = (\ccalV, \ccalE, \ccalW)$, where $\ccalV$ is the set of $N$ nodes, $\ccalE \subseteq \ccalV \times \ccalV$ is the set of edges, and $\ccalW : \ccalE \to \reals$ is the function that assigns weights to the edges. The neighborhood of node $i \in \ccalV$ is then defined as the set of nodes $\ccalN_{i} = \{j \in \ccalV : (j,i) \in \ccalE\}$. With these notations in place, a graph signal is defined as a map which associates a real value to each element of $\ccalV$. This graph signal can be conveniently represented as the vector $\bbx \in \reals^{N}$, where the $i$-th element $[\bbx]_{i} = x_{i}$ corresponding to the value of the signal at node $i$. In order to relate the values of the graph signal at any node with those at its neighborhood, we can make use of a matrix description of the graph. More precisely, let $\bbS \in \reals^{N \times N}$ be a \emph{graph shift operator} (GSO) which is a matrix whose $(i,j)$-th element can be nonzero if and only if $i=j$ or if $(j,i) \in \ccalE$ \cite{sandryhaila14-freq}. Note then, that $\bbS \bbx$ is a linear combination of the values of the signal with that of its neighbors. More precisely, we have that, for each $i \in \ccalV$
\begin{equation} \label{eqn_neighbor}
\vspace{\intereq}
[\bbS\bbx]_{i} 
	= \sum_{j = 1}^{N} [\bbS]_{ij} x_{j} 
	= \sum_{j \in \ccalN_{i} \cup \{i\}} [\bbS]_{ij} x_{j}
\vspace{\intereq}
\end{equation}
where the second equality follows because $[\bbS]_{ij}=0$ if $j \notin \ccalN_{i}\cup \{i\}$. The operation in \eqref{eqn_neighbor} is the basic element to extend the notion of convolution (filtering) to signals defined on graphs; see, e.g., \cite{segarra17-linear}. First, observe that while $\bbS \bbx$ collects information from the one-hop neighborhood of each node, $\bbS^{k} \bbx$ collects information up to the $k$-hop neighborhood of each node. Denote by $\bbh = [h_{0},\ldots,h_{K-1}]^{\Tr} \in \reals^{K}$ a collection of $K$ filter taps. Then, we can linearly combine neighboring values up to the $(K-1)$-hop neighborhood by [cf. \eqref{eqn_regular_conv}]
\begin{equation} \label{eqn_irregular_conv}
\vspace{\intereq}
[\bbh \ast \bbx]_{i} = \sum_{k=0}^{K-1} h_{k} [\bbS^{k}\bbx]_{i}.
\vspace{\intereq}
\end{equation}
Upon defining matrix $\bbH := \sum_{k=0}^{K-1} h_{k} \bbS^{k} \in \reals^{N \times N}$, \eqref{eqn_irregular_conv} can be equivalently written as
\begin{equation}
\vspace{\intereq}
\bbh \ast \bbx = \bbH \bbx = \sum_{k=0}^{K-1} h_{k} \bbS^{k} \bbx.
\vspace{\intereq}
\end{equation}
with $\bbH$ being known as a linear shift-invariant (LSI) GF \cite{segarra17-linear}.

Since LSI-GFs are regarded as the generalization of convolutions to operate on graph signals, the operator in \eqref{eqn_irregular_conv} can be used to extend CNNs to operate on graph signals \cite{bruna14-deepspectralnetworks}. More specifically, assume that in each layer $\ell$ of the CNN, output $\bbx_{\ell}$ consists of $F_{\ell}$ features, each of which is considered a graph signal $\bbx_{\ell}^{(f)} \in \reals^{N_{\ell} \times 1}$ defined on an $N_{\ell}$-node graph described by GSO $\bbS_{\ell} \in \reals^{N_{\ell} \times N_{\ell}}$, $f=1,\ldots,F_{\ell}$. Then, all $F_{\ell}$ features can be concatenated on vector $\bbx_{\ell} = [(\bbx_{\ell}^{(1)})^{\Tr},\ldots,(\bbx_{\ell}^{(F_{\ell})})^{\Tr}]^{\Tr} \in \reals^{F_{\ell}N_{\ell} \times 1}$. Assume that the linear transform $\bbA_{\ell}$ constructs $F_{\ell}$ features out of the existing $F_{\ell-1}$ ones. Then, $\bbA_{\ell}$ can be regarded as a MIMO GF since it takes $F_{\ell-1}$ input signals and outputs $F_{\ell}$ graph signals. By denoting as $\otimes$ the Kronecker matrix product, the output of the convolution operation on graph signals can be compactly written as a MIMO GF as follows
\begin{equation} \label{eqn_graph_conv}
\vspace{\intereq}
\bbA_{\ell} \bbx_{\ell-1} = \sum_{k=0}^{K_{\ell}-1} \left( \bbH_{\ell,k} \otimes \bbS_{\ell}^{k} \right) \bbx_{\ell-1}
\vspace{\intereq}
\end{equation}
where $\bbH_{\ell,k} \in \reals^{F_{\ell} \times F_{\ell-1}}$ contains the filter taps corresponding to the $F_{\ell-1} F_{\ell}$ LSI-GFs employed. More precisely, by denoting as $[\bbH_{\ell,k}]_{f,f'} = h_{k,f,f'}^{(\ell)}$, the filter taps of the $(f,f')$ filter can be written as $\bbh_{f,f'}^{(\ell)} = [h_{0,f,f'}^{(\ell)},\ldots,h_{K_{\ell-1},f,f'}^{(\ell)}]^{\Tr} \in \reals^{K_{\ell} \times 1}$, $f=1,\ldots,F_{\ell}$, $f'=1,\ldots,F_{\ell-1}$. Construction \eqref{eqn_graph_conv} builds $F_{\ell}$ different LSI-GFs for each of the $F_{\ell-1}$ features contained in the previous layer, totaling $F_{\ell-1}F_{\ell}$ LSI-GFs. The total number of \emph{trainable} parameters is thus  $F_{\ell-1}F_{\ell} K_{\ell}$. While written differently, the MIMO GF in \eqref{eqn_graph_conv} represents the per-layer architecture  proposed in \cite{defferrard17-cnngraphs}.

Finally, with respect to the pooling operation, the use of multiscale hierarchical clustering to reduce the size of the graph in each layer has been employed, yielding $N_{\ell} \leq N_{\ell-1}$ and $\bbS_{\ell}$ the corresponding GSO of each layer \cite{bruna14-deepspectralnetworks, defferrard17-cnngraphs}. Also, due to the computational and performance issues of clustering, alternative approaches which do not rely on pooling exist \cite{gama17-nvgf}. In this work, we focus on the convolutional operation of CNNs on graph signals, letting the user determine the preferred choice of pooling scheme.

\vspace{\intersection}

\section{CNNs based on structured MIMO GFs} \label{sec_simple}

This paper proposes three new architectures for CNNs on graph signals, obtained by imposing a certain parsimonious representation on the MIMO GF matrices $\{\bbH_{\ell,1},\ldots,\bbH_{\ell,K_{\ell}}\}$. The resulting architectures yield a considerably lower number of trainable parameters, reducing the complexity of the network, as well as avoiding certain pitfalls such as overfitting or the curse of dimensionality \cite{huang17-densecnn}. For simplicity, from now on, we focus on some specific layer $\ell$, hence dropping the subscript on all notations. We denote as $\bbx = \bbx_{\ell-1}$ the input, with $\bbx = [\bbx_{1}^{\Tr},\ldots,\bbx_{Q}^{\Tr}]^{\Tr} \in \reals^{QN \times 1}$ where $\bbx_{q} \in \reals^{N}$ are the $F_{\ell-1}=Q$ input features, $q=1,\ldots,Q$. We denote as $\bby = \bbx_{\ell}$ the output features, with $\bby = [\bby_{1}^{\Tr},\ldots,\bby_{P}^{\Tr}]^{\Tr} \in \reals^{PN \times 1}$ where $\bby_{p} \in \reals^{N \times 1}$ are the $F_{\ell}=P$ output features, $p=1,\ldots,P$. The length of the filters is $K_{\ell}=K$ and the matrix of filter taps $\bbH_{\ell,k} = \bbH_{k} \in \reals^{P \times Q}$ has elements $[\bbH_{k}]_{p,q} = h_{k,p,q}$, $k=0,\ldots,K-1$, $p=1,\ldots,P$, $q=1,\ldots,Q$. Each of the $(p,q)$ filters is represented by a vector of filter taps $\bbh_{p,q} = [h_{0,p,q},\ldots,h_{K-1,p,q}]^{\Tr} \in \reals^{K \times 1}$. Equation \eqref{eqn_graph_conv} becomes
\begin{equation} \label{eqn_mimo}
\vspace{\intereq}
\bby = \sum_{k=0}^{K-1} \left( \bbH_{k} \otimes \bbS^{k} \right) \bbx
\vspace{\intereq}
\end{equation}
and each new feature is computed as
\begin{equation} \label{eqn_yp}
\vspace{\intereq}
\bby_{p} = \sum_{q=1}^{Q} \sum_{k=0}^{K-1} h_{k,p,q} \bbS^{k} \bbx_{q} \ , \ p=1,\ldots,P.
\vspace{\intereq}
\end{equation}
The design variables are the collection of matrices $\{\bbH_{0},\ldots,\bbH_{K-1}\}$ containing the $PQ$ filters, and thus totaling $PQK$ parameters.

\vspace{\intersection}


\subsection{Aggregating the input features} \label{subsec_P}

First, we propose to aggregate all input features so as to reduce the number of filters. Instead of designing $P$ different filters for each one of the $Q$ input features, we first aggregate the $Q$ input features into one graph signal, and proceed to design $P$ different filters to be applied to this graph signal.

This strategy amounts to designing filter taps $\bbh_{p,1} = [h_{0,p,1},\ldots,h_{K-1,p,1}]^{\Tr} \in \reals^{K \times 1}$ for $p=1,\ldots,P$. Then, matrix $\bbH_{k}$ in \eqref{eqn_mimo} becomes a replication of the first column
\begin{equation} \label{eqn_Hk_P}
\vspace{\intereq}
\bbH_{k} = 
	\begin{bmatrix}
		h_{k,1,1} & h_{k,1,1} & \cdots & h_{k,1,1} \\
		h_{k,2,1} & h_{k,2,1} & \cdots & h_{k,2,1} \\
		\vdots    & \vdots    & \ddots & \vdots    \\
		h_{k,P,1} & h_{k,P,1} & \cdots & h_{k,P,1}
	\end{bmatrix}.
\vspace{\intereq}
\end{equation}
Each output feature \eqref{eqn_yp} is thus computed as
\begin{equation} \label{eqn_yp_P}
\bby_{p} = \sum_{q=1}^{Q} \sum_{k=1}^{K} h_{k,p,1} \bbS^{k} \bbx_{q} = \sum_{k=1}^{K} h_{k,p,1} \bbS^{k} \left( \sum_{q=1}^{Q} \bbx_{q} \right)
\end{equation}
for $p=1,\ldots,P$.

Observe that the structure imposed on \eqref{eqn_Hk_P} leads to only $P$ different LSI-GFs and therefore the number of trainable parameters has been reduced to $PK$. The effect of this filter, as observed from \eqref{eqn_yp_P} is to first aggregate all the input features into a single graph signal $\sum_{q=1}^{Q} \bbx_{q}$ and then applying $P$ different filters to it, yielding the $P$ different output features.

\vspace{\intersection}


\subsection{Consolidating output features} \label{subsec_Q}

The second proposed architecture consists of designing one filter for each input feature and then consolidating all the filtered input features into a single output feature.

In order to do this, we need to design $Q$ LSI-GFs described by filter taps $\bbh_{1,q} = [h_{0,1,q},\ldots,h_{K-1,1,q}]^{\Tr} \in \reals^{K-1}$ specific to each input feature $q=1,\ldots,Q$. Matrix $\bbH_{k}$ in \eqref{eqn_mimo} can thus be written as a replication of the first row
\begin{equation} \label{eqn_Hk_Q}
\vspace{\intereq}
\bbH_{k} = 
	\begin{bmatrix}
		h_{k,1,1} & h_{k,1,2} & \cdots & h_{k,1,Q} \\
		h_{k,1,1} & h_{k,1,2} & \cdots & h_{k,1,Q} \\
		\vdots    & \vdots    & \ddots & \vdots    \\
		h_{k,1,1} & h_{k,1,2} & \cdots & h_{k,1,Q}
	\end{bmatrix}.
\vspace{\intereq}
\end{equation}
This leads to each output feature \eqref{eqn_yp} being calculated as
\begin{equation} \label{eqn_yp_Q}
\vspace{\intereq}
\bby_{p} = \sum_{q=1}^{Q} \left( \sum_{k=0}^{K-1} h_{k,1,q} \bbS^{k} \right) \bbx_{q} = \sum_{q=1}^{Q} \bbG_{q} \bbx_{q}
\vspace{\intereq}
\end{equation}
for $p=1,\ldots,P$.

Note that all $P$ output features $\bby_{p}$ are actually the same, since \eqref{eqn_yp_Q} does not depend on $p$. Thus, the filter taps given by \eqref{eqn_Hk_Q} actually yield a single output feature, which is the graph signal given by \eqref{eqn_yp_Q}. This could be particularly useful for reducing operational complexity of subsequent layers, since graph signals can be handled easily. Observe that the imposed structure \eqref{eqn_Hk_Q} containing $Q$ distinct filters yields $QK$ trainable parameters.


\begin{figure*}
\centering
\begin{subfigure}{.33\textwidth}
  \centering
  \includegraphics[width=0.9\textwidth]{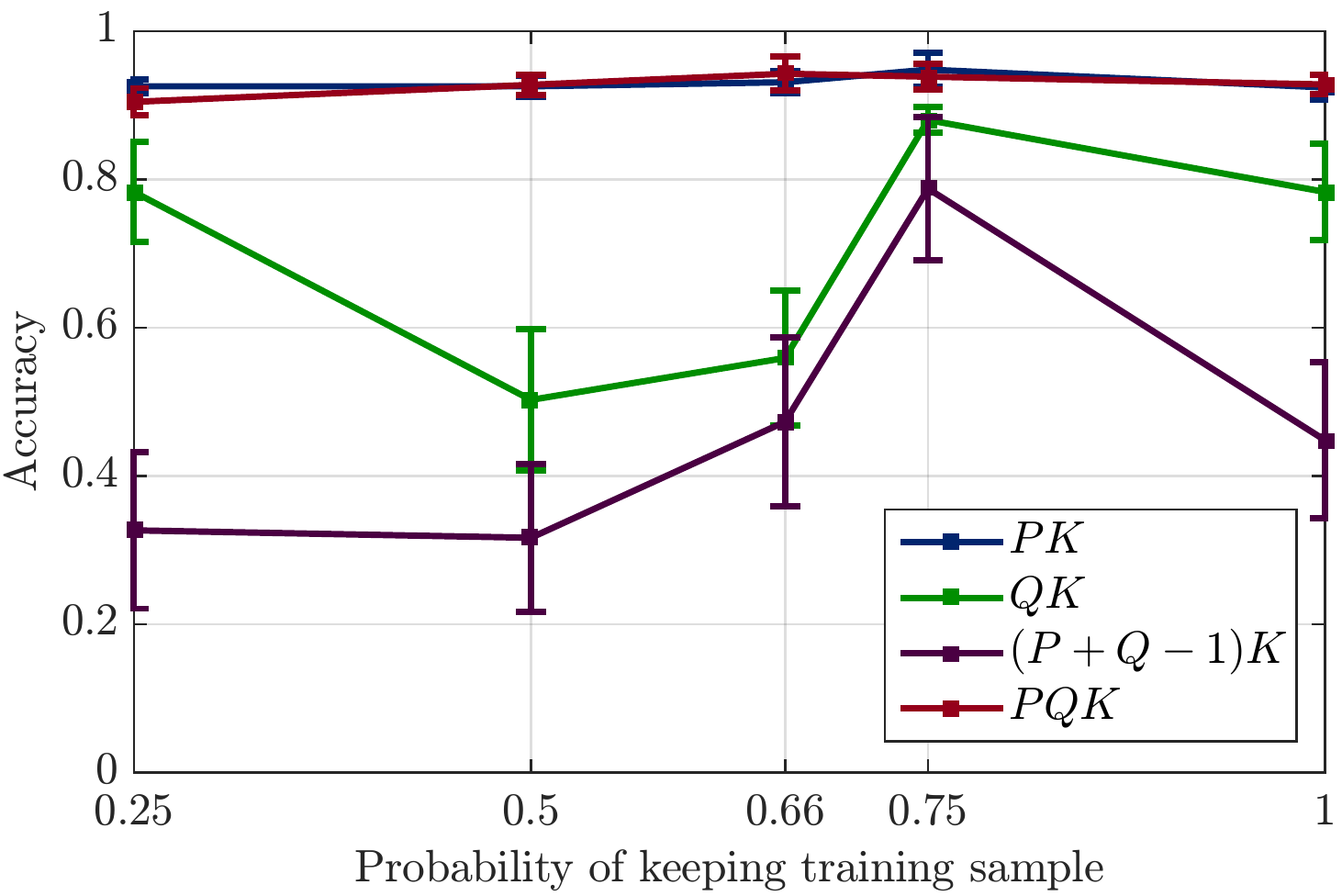}
  \caption{}
  \label{fig_keep_prob}
\end{subfigure}%
\begin{subfigure}{.33\textwidth}
  \centering
  \includegraphics[width=0.9\textwidth]{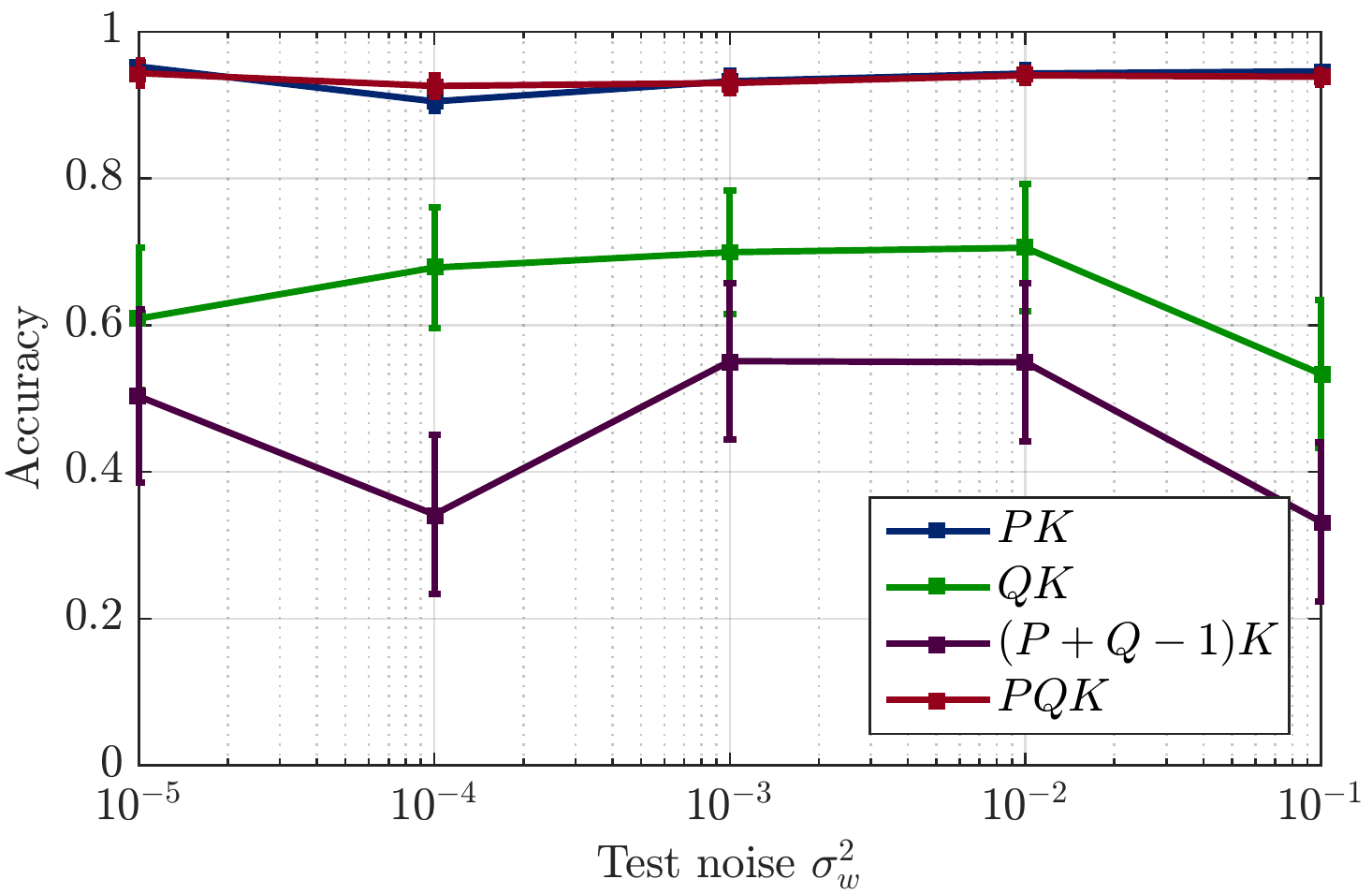}
  \caption{}
  \label{fig_sigma2}
\end{subfigure}%
\begin{subfigure}{.33\textwidth}
  \centering
  \includegraphics[width=0.8\textwidth]{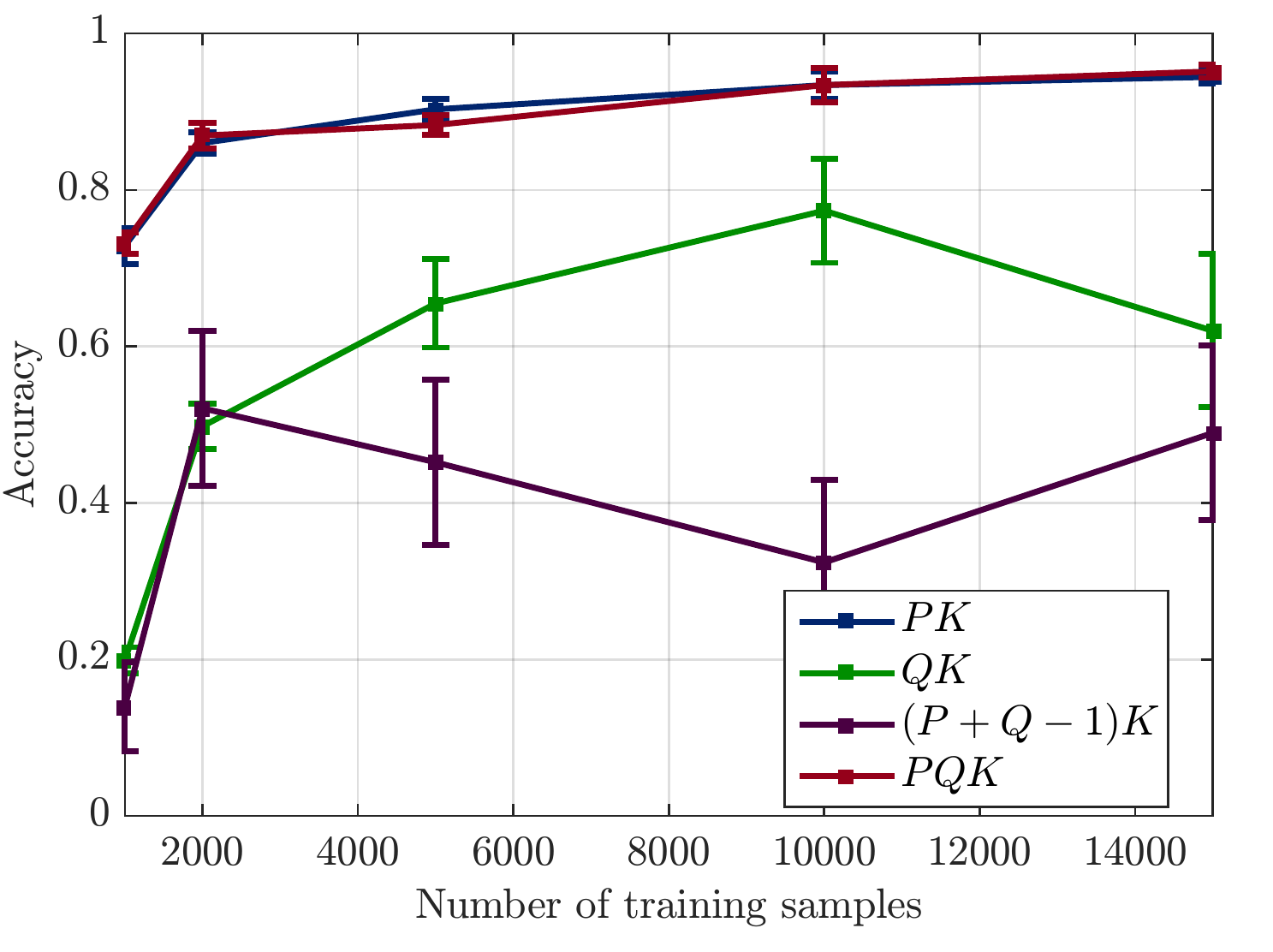}
  \caption{}
  \label{fig_ntrain}
\end{subfigure}%

\vspace{-0.1in}

\caption{Accuracy in the source localization problem. Results were averaged across $10$ different graph realizations. For clarity of figures, error bars represent $1/4$ of the estimated variance. (\subref{fig_keep_prob}) As a function of the probability of keeping a sample during the training phase, i.e. $1-\textrm{prob}_{\textrm{dropout}}$. (\subref{fig_sigma2}) As a function of the noise in the test set. (\subref{fig_ntrain}) As a function of the number of training samples. Overall, we observe that the full architecture \eqref{eqn_graph_conv} yields a performance similar to that proposed in Section~\ref{subsec_P}.}
\vspace{-0.05in}
\label{fig_sourceloc}
\end{figure*}

\vspace{\intersection}

\subsection{Convolution of features} \label{subsec_PQ}

As a third approach to reducing the number of parameters involved in \eqref{eqn_mimo}, we consider a set of $P+Q-1$ filters to be applied sequentially and progressively to the input features, yielding output features that resemble convolutions of the input features.

Let $\bbh_{p-q+1}=[h_{0,p-q+1},\ldots,h_{K-1,p-q+1}]^{\Tr} \in \reals^{K-1}$ be a set of filter taps, $p=1,\ldots,P$, $q=1,\ldots,Q$. Then, for this third proposed strategy, matrix $\bbH_{k}$ in \eqref{eqn_mimo} becomes
\begin{equation} \label{eqn_Hk_PQ}
\vspace{\intereq}
\bbH_{k} = 
	\begin{bmatrix}
		h_{k,1} & h_{k,0}   & \cdots & h_{k,2-Q}   \\
		h_{k,2} & h_{k,1}   & \cdots & h_{k,3-Q}   \\
		\vdots  & \vdots    & \ddots & \vdots      \\
		h_{k,P} & h_{k,P-1} & \cdots & h_{k,P+1-Q} \\
	\end{bmatrix}.
\vspace{\intereq}
\end{equation}
The output features are then obtained using \eqref{eqn_Hk_PQ} in \eqref{eqn_yp} yielding
\begin{equation} \label{eqn_yp_PQ}
\vspace{\intereq}
\bby_{p} 
	= \sum_{q=1}^{Q} \sum_{k=0}^{K-1} h_{k,p+1-q} \bbS^{k} \bbx_{q}
	= \sum_{k=0}^{K-1} \bbS^{k} \sum_{q=1}^{Q} h_{k,p+1-q} \bbx_{q}
\vspace{\intereq}
\end{equation}
for $p=1,\ldots,P$.

From \eqref{eqn_yp_PQ} we observe that each output feature $\bby_{p}$ can be thought of as the convolution of the input feature vector $\bbx$ with the collection of filter taps given by $\{h_{k,p},\ldots,h_{k,p+1-Q}\}$. Note also that consecutive output features weigh input features similarly. In a way, \eqref{eqn_yp_PQ} acts as a smoother of input features. Matrix $\bbH_{k}$ in \eqref{eqn_Hk_PQ} is a Toeplitz matrix and thus has $P+Q-1$ elements, so that the total number of trainable parameters is $(P+Q-1)K$.

\vspace{\intersection}

\section{Numerical tests} \label{sec_sims}

Here we compare the performance of the three architectures proposed in Sections~\ref{subsec_P},~\ref{subsec_Q}~and~\ref{subsec_PQ}, which involve, respectively, $PK$, $QK$ and $(P+Q-1)K$ parameters, with that of \cite{defferrard17-cnngraphs}, which involves $PQK$ parameters [cf. \eqref{eqn_graph_conv}]. 
In the first testcase we consider a synthetic dataset of a source localization problem in which different diffused graph signals are processed to determine the single node that originated them. In the second testcase we use the \texttt{20NEWS} dataset and a \texttt{word2vec} embedding underlying graph to classify articles in one out of $20$ different categories \cite{joachims96-20news}. For both problems, we evaluate an architecture with $2$ convolutional layers, the first one generating $F_{1}=32$ features, and the second one outputting $F_{2}=64$ features. GFs are of length $K_{1}=K_{2}=K=5$. No pooling is employed, so that $N_{1}=N_{2}=N$, the number of nodes in the graph, and $\bbS_{1}=\bbS_{2}=\bbS$ is the GSO specific to each testcase. We denote as $PQK$ the architecture in \cite{defferrard17-cnngraphs}, and as $PK$, $QK$ and $(P+Q-1)K$ the ones developed in Sections~\ref{subsec_P},~\ref{subsec_Q}~and~\ref{subsec_PQ}, respectively. The number of parameters in the convolutional layers are $10400$ for $PQK$, $480$ for $PK$, $165$ for $QK$ and $635$ for $(P+Q-1)K$. The selected nonlinearity is a ReLU applied at each layer and all architectures include a readout layer. For the training stage in both problems, an ADAM optimizer with learning rate $0.005$ was employed \cite{kingma17-adam}, for $20$ epochs and batch size of $100$.

\begin{table}
	\centering
\begin{tabular}{lrc} \hline
Architecture 	& Parameters 	& Accuracy 	\\ \hline
$PQK$			& $10,400$		& $93.9\%$	\\ 
$PK$			& $   480$		& $94.8\%$	\\
$QK$			& $   165$		& $88.0\%$	\\
$(P+Q-1)K$		& $   635$		& $78.8\%$	\\ \hline
\end{tabular}
 \vspace{-0.05in}
	\caption{Source localization results for $N=16$ nodes.}
	\vspace{-0.15in}
	\label{table_sourceloc}
\end{table}

\vspace{\intersection}


\subsection{Source localization.} \label{subsec_source}
Consider a connected Stochastic Block Model (SBM) graph with $N=16$ nodes divided in $4$ communities, with intracommunity edge probability of $0.8$ and intercommunity edge probability of $0.2$. Let $\bbW$ denote its adjacency matrix. With $\bbdelta_{c}$ representing a graph signal taking the value $1$ at node $c$ and $0$ elsewhere, the signal $\bbx = \bbW^{t} \bbdelta_{c}$ is a diffused version of the sparse input $\bbdelta_{c}$ for some unknown $0 \leq t \leq N-1$. The objective is to determine the source $c$ that originated the signal $\bbx$ irrespective of time $t$. To that end, we create a set of $N_{\textrm{train}}$ labeled training samples $\{(c',\bbx')\}$ where $\bbx' = \bbW^{t} \bbdelta_{c'}$ with both $c'$ and $t$ chosen at random. Then we create a test set with $N_{\textrm{test}}$ samples in the same fashion, but we add i.i.d. zero-mean Gaussian noise $\bbw$ with variance $\sigma_{w}^{2}$, so that the signals to be classified are $ \bbW^{t} \bbdelta_{c} + \bbw$. The goal is to use the training samples to design a CNN that determines the source (node) $c$ that originated the diffused signal.

First, we run the source localization problem on $10$ different realizations of randomly generated SBM graphs. For each graph, we train the four architectures using dropout with probability of keeping each training sample of $0.75$. The total number of training samples is $10,000$. Once trained, the architectures are tested on a test set of $200$ samples for each graph, contaminated with noise of variance $\sigma_{w}^{2} = 10^{-1}$. Results are listed in Table~\ref{table_sourceloc}, where the accuracy is averaged over the $200$ samples, over the $10$ different graph realizations. We observe that the $PK$ architecture proposed in Section~\ref{subsec_P} outperforms the full $PQK$ architecture, with two orders of magnitude less parameters. Also, the $QK$ and the $(P+Q-1)K$ architectures yields reasonable performances.

\begin{table}
	\centering
\begin{tabular}{lrc} \hline
Architecture 	& Parameters 	& Accuracy 	\\ \hline
$PQK$			& $10,400$		& $61.32\%$	\\ 
$PK$			& $   480$		& $62.48\%$	\\
$QK$			& $   165$		& $58.22\%$	\\
$(P+Q-1)K$		& $   635$		& $64.06\%$	\\ \hline
\end{tabular}
 \vspace{-0.05in}
	\caption{Results for classification on \texttt{20NEWS} dataset on a \texttt{word2vec} graph embedding of $N=5,000$ nodes.}
	\vspace{-0.15in}
	\label{table_20news}
\end{table}

Additionally, we run tests changing the values of several of the simulations parameters. In Fig.~\ref{fig_keep_prob} we observe the accuracy obtained when varying the probability of keeping training samples. It is noted that the $PK$ architecture performs as well as the full $PQK$ architecture. It is also observed that the other two architectures have significant variance, which implies that they depend heavily on the topology of the graph. The effect of noise $\sigma_{w}^{2}$ on the test samples can be observed in Fig.~\ref{fig_sigma2}. We observe that all four architectures are relatively robust to noise. The $QK$ and $(P+Q-1)K$ architectures exhibit a dip in performance for the highest noise value simulated. Finally, in Fig.~\ref{fig_ntrain} we show the performance of all four architectures as a function of the number of training samples. We see that the $PK$ and the full $PQK$ architecture improve in performance as more training samples are considered, with a huge increase between $1,000$ and $2,000$ training samples. This same increase is observed for the remaining two architectures, although their performance behaves somewhat erratically afterwards. All in all, from these set of simulations, we observe that the $PK$ architecture performs as well as the full $PQK$ architecture, but utilizing almost two orders of magnitude less parameters. We also observe that the $QK$ and $(P+Q-1)K$ architectures have a high dependence on the topology of the network.

\vspace{\intersection}


\subsection{\texttt{20NEWS} dataset} \label{subsec_20news}

Here we consider the classification of articles in the \texttt{20NEWS} dataset which consists of $18,846$ texts ($11,314$ of which are used for training and $7,532$ for testing) \cite{joachims96-20news}. The graph signals are constructed as in \cite{defferrard17-cnngraphs}: each document $x$ is represented using a normalized bag-of-words model and the underlying graph support is constructed using a $16$-NN graph on the \texttt{word2vec} embedding \cite{mikolov13-word2vec} considering the $5,000$ most common words. The GSO adopted is the normalized Laplacian. No dropout is used in the training phase. Accuracy results are listed in Table~\ref{table_20news}, demonstrating that the $PK$ and $(P+Q-1)K$ architectures outperform the full $PQK$ one, but requiring almost $100$ times less parameters.

\vspace{\intersection}

\section{Conclusions} \label{sec_conclusions}

In this paper, we have studied the problem of extending CNNs to operate on graph signals. More precisely, we reframed existing architectures under the concept of MIMO GFs, and leveraged structured representations of such filters to reduce the number of trainable parameters involved. We proposed three new architectures, each of which arises from adopting a different parsimonious model on the MIMO GF matrices. All the resulting architectures yield a lower number of trainable parameters, reducing computational complexity, as well as helping in avoiding certain pitfalls of training like overfitting or the curse of dimensionality.

We have applied the three proposed architectures to a synthetic problem on source localization, and compared its performance with the more complex, full MIMO GF model. We analyzed performance as a function of dropout probability in the training phase, noise in the test samples, and number of training samples. We noted that the proposed architecture that aggregates input features (Section~\ref{subsec_P}) has a performance similar to that of the full model, but involving two orders of magnitude less parameters. The other two architectures offer comparable performance for certain values of the analyzed parameters. Finally, we utilized the proposed architectures on the problem of classifying articles of the \texttt{20NEWS} dataset. In this case, we observed that two of the proposed parsimonious models outperform the full model.


\bibliographystyle{IEEEbib}
\bibliography{myIEEEabrv,bib-mimo}

\end{document}